\documentclass[twoside]{article}
\newif\ifisarxiv
\isarxivtrue

\ifisarxiv
\usepackage[accepted]{sty/ridge_arxiv}
\else
\usepackage[accepted]{sty/aistats2018_style/aistats2018}
\fi

\usepackage{amsmath}
\usepackage{amssymb}
\usepackage{color}
\usepackage{cancel}
\usepackage{algorithm}
\usepackage{algorithmic}
\usepackage{graphicx}

\def\xib{\boldsymbol\xi}

\newcommand{\ofsub}[1]{\mbox{\small \raisebox{0.0pt}{$(#1)$}}}
\newcommand{\of}[2]{{#1{\!\ofsub{#2}}}}

\newcommand{\lazy}{FastRegVol}
\newcommand{\volsamp}{RegVol}

\newcommand{\Sm}{{S_{-i}}}

\ifx\BlackBox\undefined
\newcommand{\BlackBox}{\rule{1.5ex}{1.5ex}}  % end of proof
\fi

\DeclareMathOperator*{\argmin}{\mathop{\mathrm{argmin}}}
\def\x{\mathbf x}
\def\y{\mathbf y}

\def\w{\mathbf w}
\def\v{\mathbf v}
\def\wbh{\widehat{\mathbf w}}
\def\e{\mathbf e}

\def\X{\mathbf X}

\def\Z{\mathbf Z}
\def\I{\mathbf I}

\def\Rh{\widehat{R}}
\def\Ot{\widetilde{O}}

\def\E{\mathbb E}
\def\R{\mathbb R} 
\def\tr{\mathrm{tr}}

\def\Var{\mathrm{Var}}

\newcommand{\defeq}{\stackrel{\text{\tiny{def}}}{=}}

\definecolor{silver}{cmyk}{0,0,0,0.3}
\definecolor{yellow}{cmyk}{0,0,0.9,0.0}
\definecolor{reddishyellow}{cmyk}{0,0.22,1.0,0.0}
\definecolor{black}{cmyk}{0,0,0.0,1.0}
\definecolor{darkYellow}{cmyk}{0.2,0.4,1.0,0}
\definecolor{darkSilver}{cmyk}{0,0,0,0.1}
\definecolor{grey}{cmyk}{0,0,0,0.5}
\definecolor{darkgreen}{cmyk}{0.6,0,0.8,0}

\newcommand{\Green}[1]{{\color{darkgreen}  {#1}}}
\newcommand{\Blue}[1]{\color{blue}{#1}\color{black}}
\newcommand{\Brown}[1]{{\color{brown}{#1}\color{black}}}

\ifx\proof\undefined
\newenvironment{proof}{\par\noindent{\bf Proof\ }}{\hfill\BlackBox\\[2mm]}
\fi

\ifx\theorem\undefined
\newtheorem{theorem}{Theorem}
\fi

\ifx\example\undefined
\newtheorem{example}{Example}
\fi

\ifx\property\undefined
\newtheorem{property}{Property}
\fi

\ifx\lemma\undefined
\newtheorem{lemma}[theorem]{Lemma}
\fi

\ifx\proposition\undefined
\newtheorem{proposition}[theorem]{Proposition}
\fi

\ifx\remark\undefined
\newtheorem{remark}[theorem]{Remark}
\fi

\ifx\corollary\undefined
\newtheorem{corollary}[theorem]{Corollary}
\fi

\ifx\definition\undefined
\newtheorem{definition}{Definition}
\fi

\ifx\conjecture\undefined
\newtheorem{conjecture}[theorem]{Conjecture}
\fi

\ifx\axiom\undefined
\newtheorem{axiom}[theorem]{Axiom}
\fi

\ifx\claim\undefined
\newtheorem{claim}[theorem]{Claim}
\fi

\ifx\assumption\undefined
\newtheorem{assumption}[theorem]{Assumption}
\fi

\begin{document}

\runningtitle{Subsampling for Ridge Regression
via Regularized Volume Sampling}


\twocolumn[

\aistatstitle{Subsampling for Ridge Regression\\
via Regularized Volume Sampling$^*$}

\aistatsauthor{ Micha{\l } Derezi\'{n}ski \And Manfred K. Warmuth}

\aistatsaddress{
Department of Computer Science\\
University of California Santa Cruz\\
\texttt{mderezin@ucsc.edu}\And 
Department of Computer Science\\
University of California Santa Cruz\\
\texttt{manfred@ucsc.edu}} ]

\begin{abstract}
Given $n$ vectors $\x_i\in\R^d$, we want to fit a
linear regression model for noisy labels $y_i\in\R$. 
The ridge estimator is a classical solution to this problem. 
However, when labels are expensive, 
we are forced to select only a small subset of vectors $\x_i$
for which we obtain the labels $y_i$. We propose a new
procedure for selecting the subset of vectors, such that the ridge
estimator obtained from that subset offers strong statistical
guarantees in terms of the mean squared prediction error over the
entire dataset of $n$ labeled vectors. 
The number of labels needed is proportional to the statistical
dimension of the problem which is often much smaller than $d$. 
Our method is an extension of a joint subsampling procedure called
volume sampling. A second major contribution is that we speed up volume
sampling so that it is essentially as efficient as leverage
scores, which is the main i.i.d. subsampling procedure for this task.
Finally, we show theoretically and experimentally that volume sampling
has a clear advantage over any i.i.d. sampling when labels are expensive.
\end{abstract}

\section{Introduction}
\label{sec:introduction}
{\renewcommand{\thefootnote}{\fnsymbol{footnote}}
\footnotetext[1]{Supported by NSF grant IIS-1619271.}}

Given a matrix $\X\in \R^{d\times n}$, we consider the
task of fitting a linear model\footnote{This setting can easily be
  extended to ``non-linear models'' via kernelization.} to a vector of labels
$\y=\X^\top\w^*+\xib$, where $\w^*\in\R^d$ and the noise
$\xib\in\R^n$ is a
mean zero random vector with covariance matrix $\Var[\xib]\preceq\sigma^2\I$
for some $\sigma>0$. A classical solution to this task is the ridge estimator:
\begin{align*}
\wbh_\lambda^* &= 
\argmin_{\w\in\R^d} \;\; 
\|\X^\top\w - \y\|^2 + \lambda\|\w\|^2
\\&=(\X\X^\top+\lambda\I)^{-1}\X\y.
\end{align*}
In many settings, obtaining labels $y_i$ is expensive and we are
forced to select a subset $S\subseteq\{1..n\}$ of label indices.
Let $\y_S\in \R^{|S|\times 1}$ be the sub vector of labels indexed by $S$
and $\X_S\in \R^{d\times |S|}$ be the columns of $\X$ indexed by $S$.
We will show that if $S$ is sampled with a new variant of {\em volume
  sampling} \cite{avron-boutsidis13}
on the columns of $\X$, 
then the ridge estimator for the subproblem $(\X_S,\y_S)$
$$
\of{\wbh_\lambda^*} S = (\X_S\X_S^\top+\lambda\I)^{-1}\X_S\y_S 
$$
has strong generalization properties with respect to the
full problem $(\X,\y)$.

Volume sampling is a sampling technique which has received
a lot of attention recently \cite{avron-boutsidis13, unbiased-estimates, efficient-volume-sampling,
 pca-volume-sampling, dual-volume-sampling}.
For a fixed size $s\ge d$, the original variant samples $S\subseteq \{1..n\}$  
of size $s$ proportional to the squared volume of the
parallelepiped spanned by the rows of $\X_S$
\cite{avron-boutsidis13}:
\begin{align}
P(S) \propto \det(\X_S\X_S^\top).\label{eq:probability}
\end{align}
A simple approach for implementing volume sampling 
(just introduced in \cite{unbiased-estimates})
is to start with the full set of column indices
$S=\{1..n\}$ and then (in reverse order) select an index
$i$ in each iteration to be
eliminated from set $S$ with probability proportional to the change in
matrix volume caused by removing the $i$th column: 
\begin{align}
&\text{Sample }i \sim P(i\,|\,S)  =
  \frac{\det(\X_\Sm\X_\Sm^\top)}{(|S|-d) \det(\X_S\X_S^\top)},
\label{eq:reviter}\\
&\text{Update }S \leftarrow S - \{i\}.\nonumber\\
\tag{\bf Reverse Iterative Volume Sampling}
\end{align}
Note that when $|S|<d$, then all matrices
$\X_S\X_S^\top$ are singular, and so the distribution becomes
undefined. Motivated by this limitation, we
propose a regularized variant, called $\lambda$-regularized
volume sampling:
%\footnote{The original volume sampling is retained as the special
%case when $\lambda=0$ and $s\ge d$.
%Note that for $\lambda>0$, this procedure is not the same as sampling
%$S$ st $P(S)\propto \det(\X_S\X_S^\top+\lambda\I)$
%because the products of normalization factors along paths
%from $\{1..n\}$ to $S$ are path dependent.}:
%\begin{algorithm}[H] 
% {\fontsize{8}{8}\selectfont
% \caption{\small $\lambda$-Regularized Volume Sampling}
%\label{alg:reverse-iterative}
\begin{align}
&\text{Sample }i \sim P(i\,|\,S)  \propto
  \frac{\det(\X_\Sm\X_\Sm^\top+\lambda\I)}{\det(\X_S\X_S^\top+\lambda\I)},
\label{eq:regularized}\\
&\text{Update }S \leftarrow S - \{i\}.\nonumber\\
\tag{\bf $\lambda$-Regularized Volume Sampling}
\end{align}
Note that in the special case of no regularization (i.e.
$\lambda=0$), then \eqref{eq:regularized} sums to $\frac{1}{|S|-d}$ 
(see equality \eqref{eq:reviter}).
However when $\lambda>0$, then the sum of
\eqref{eq:regularized} depends on all columns of $\X_S$ and not just
the size of $S$.
This makes regularized volume sampling more complicated and 
certain equalities proven in \cite{unbiased-estimates}
for $\lambda=0$ become inequalities.

Nevertheless, we were able to show that the proposed 
$\lambda$-regularized distribution 
exhibits a fundamental connection to ridge
regression, and introduce an efficient algorithm to sample from
it in time $O((n+d)d^2)$. In particular, we prove that
when $S$ is sampled according to $\lambda$-regularized volume
sampling with $\lambda\leq \frac{\sigma^2}{\|\w^*\|^2}$, then the mean
squared prediction error (MSPE)  of estimator $\of{\wbh_\lambda^*}S$
over the entire dataset $\X$ is bounded:
\begin{align}
\E_S\E_{\xib}\;\frac{1}{n}\|\X^\top&(\of{\wbh_\lambda^*}S-\w^*)\|^2\leq\frac{\sigma^2
   d_\lambda}{s-d_\lambda+1},\nonumber\\
\text{where}\quad d_\lambda
  &=\tr(\X^\top(\X\X^\top\!\!+\lambda\I)^{-1}\X)\label{eq:statistical-dimension}
%=d-\lambda\tr((\X\X^\top\!\! + \lambda\I)^{-1})
\end{align}
is the statistical dimension.
If $\lambda_i$ are the eigenvalues
of $\X\X^\top$, then $d_\lambda=\sum_{i=1}^d
\frac{\lambda_i}{\lambda_i+\lambda}$.
Note that $d_\lambda$ is decreasing with $\lambda$ and $d_0=d$.
If the spectrum of the matrix $\X\X^\top$ decreases quickly
then $d_\lambda$ does so as well with increasing $\lambda$.
When $\lambda$ is properly tuned then $d_\lambda$ is
the effective degrees of freedom of $\X$. 
Our new lower bounds will show
that the above upper bound for regularized volume sampling 
is essentially optimal with respect to the choice of a subsampling procedure.

Volume sampling can be viewed as a non-i.i.d. extension of leverage score
sampling \cite{fast-leverage-scores}, a widely used method where columns are
sampled independently according to their leverage scores. 
% The
% connection between the two methods is as follows: the
% probability of a given column index $i$ being selected in set $S$
% using size $s=d$ volume sampling equals its leverage score:
% \[P(i\in S) = \x_i^\top(\X\X^\top)^{-1}\x_i.\]
Volume sampling has been shown to return better column subsets
than its i.i.d. counterpart in many applications like experimental
design, linear regression and graph theory
\cite{avron-boutsidis13,unbiased-estimates}. In this paper we
additionally show that any i.i.d. subsampling with respect to any
fixed distribution such as leverage score sampling can require
$\Omega(d_\lambda\ln(d_\lambda))$ labels to achieve any
generalization for ridge regression, compared to $O(d_\lambda)$
for regularized volume sampling. We reinforce this claim
experimentally in Section \ref{sec:experiments}.
 
The main obstacle against using volume sampling in practice
has been high computational cost. Only recently, the first
polynomial time algorithms have been proposed for exact
\cite{unbiased-estimates} and approximate \cite{dual-volume-sampling}
volume sampling (see Table \ref{tab:runtime} for comparison).
\begin{table}
\begin{center}
\begin{tabular}{c|c|c}
 & Exact & Approximate\\
\hline
\cite{dual-volume-sampling}& $O(n^4s)$&$\Ot(nd^2s^3)$\\
\cite{unbiased-estimates}& $O(n^2d)$&-\\
\textbf{here}&$O(nd^2)$ & -%$\Ot(nd^2)$
\end{tabular}
\caption{Comparison of runtime for exact and approximate volume
  sampling algorithms, where $d\leq s\leq n$.}  
\label{tab:runtime}
\vspace{-4mm}
\end{center}
\end{table}
%\vspace{-3mm}
In particular, the fastest algorithm for exact volume
sampling\footnote{The exact time complexity is $O((n-s+d)nd)$ 
which is $O(n^2d)$ for $s<n/2$.}
is $O(n^2 d)$
whereas exact leverage score sampling\footnote{Approximate leverage
score sampling methods achieve even better runtime of $\Ot(nd+d^3)$.}
is $O(nd^2)$ (in both cases, the dependence on sample size $s$ is not
asymptotically significant). In typical settings for experimental
design \cite{optimal-design-book} and active learning
\cite{pool-based-active-learning-regression}, quality of the
sample is more important than the runtime. 
However for many modern datasets,
the number of examples $n$ is much larger than $d$,
which makes existing algorithms for volume sampling 
infeasible. In this paper, we give an easy-to-implement volume
sampling algorithm that runs in time $O(nd^2)$.
Thus we give the first volume sampling procedure which 
is essentially linear in $n$
and matches the time complexity of exact leverage score sampling.
For example, dataset MSD from the
UCI data repository \cite{uci-repository} has $n=464,000$
examples with dimension $d=90$. Our algorithm performed volume
sampling on this dataset in 39 seconds, whereas the previously best $O(n^2d)$
algorithm \cite{unbiased-estimates} did not finish within $24$ hours. 
Sampling with leverage scores took 12 seconds on this data set.
Finally our procedure also achieves regularized volume sampling
for any $\lambda > 0$ with the running time of
$O((n+d)d^2)$.

\subsection{Related work}

Many variants of probability distributions based on the matrix
determinant have been studied in the literature, including
Determinantal Point Processes (DPP) \cite{dpp}, k-DPP's
\cite{k-dpp} and volume sampling \cite{avron-boutsidis13, unbiased-estimates, efficient-volume-sampling, dual-volume-sampling}, with
applications to matrix 
approximation \cite{pca-volume-sampling}, 
%clustering \cite{dpp-clustering}, 
recommender systems \cite{dpp-shopping}, etc.  More recently,
further theoretical results suggesting applications of volume
sampling in linear regression were given by
\cite{unbiased-estimates}, where an expected loss bound for the
unregularized least squares estimator was given under volume sampling
of size $s=d$. Moreover, Reverse Iterative Volume
Sampling -- a technique enhanced in this paper with a regularization --
was first proposed in \cite{unbiased-estimates}. 
%for $\lambda=0$.

Subset selection techniques for regression have long been studied in
the field of experimental design \cite{optimal-design-book}. More
recently, computationally tractable techniques have been explored \cite{tractable-experimental-design}. Statistical guarantees under
i.i.d. subsampling in kernel ridge regression have been analyzed for uniform
sampling \cite{ridge-uniform} and leverage score sampling
\cite{ridge-leverage-scores}. 
In this paper, we propose the first
tractable non-i.i.d. subsampling procedure with strong statistical
guarantees for the ridge estimator and show its benefits over using i.i.d.
sampling approaches. 

% The particular setting
% we are considering - volume sampling of sets of fixed size $s\geq d$ - 
% was first introduced in \cite{avron-boutsidis13}, with several
% theoretical results indicating applications in linear regression and
% graph theory, but no polynomial time algorithm.
For the special case of volume sampling size $s=d$, a polynomial time
algorithm was developed by \cite{efficient-volume-sampling}, and
slightly improved by \cite{more-efficient-volume-sampling}, with
runtime $O(nd^3)$. An exact sampling algorithm for arbitrary $s\geq d$
volume sampling was given by \cite{unbiased-estimates}, with
time complexity $O((n-s+d)nd)$ which is $O(n^2d)$ when $s<n/2$. Also,
\cite{dual-volume-sampling} proposed a Markov-chain procedure which generates
$\epsilon$-approximate volume samples in time $\Ot(nd^2s^3)$. The
algorithm proposed in this paper, running in time $O(nd^2)$, enjoys
a direct asymptotic speed-up over all of the above methods. Moreover,
the procedure suffers only a small constant factor overhead over computing
exact leverage scores of matrix $\X$. 

% Two main properties make volume sampling appealing - its ability
% to select informative instances from data, and the fact that it
% encourages diversity in the sampled sets (eg, two identical columns are
% unlikely to be sampled together). Leverage score sampling, a closely
% related method which has found wide application in approximation
% methods for linear algebra \cite{fast-leverage-scores,
%   randomized-matrix-algorithms,iterative-row-sampling}, effectively
% selects informative examples, but does not favor diversity of the
% samples. The advantage of leverage score sampling has been cheaper
% computational cost of obtaining exact ($O(nd^2)$ time) or approximate samples
% ($\Ot(nd)$ time) from this distribution, which is particularly important for
% applications on large datasets, but less crucial for experimental
% design \cite{optimal-design-book} and active learning
% \cite{pool-based-active-learning-regression} settings, where quality of the
% sample is more important than the runtime. For many
% modern datasets, none of the existing volume sampling algorithms have been
% computationally viable, leaving leverage score sampling as the
% necessary alternative. In this paper we seek to close this gap by
% offering a volume sampling algorithm with runtime that has only a
% small constant factor overhead over computing leverage scores.

\subsection{Main results}

The main contributions of this paper are two-fold:
\begin{enumerate}
\item \textbf{Statistical}: We define a regularized variant of volume
  sampling and show that it offers strong generalization guarantees
  for ridge regression in terms of mean squared error (MSE) and mean
  squared prediction error (MSPE).
\item \textbf{Algorithmic}: We propose a simple implementation of
  volume sampling, which not only extends the procedure
  to its regularized variant, but also offers a significant runtime
  improvement over the existing methods when $n\gg d$.
\end{enumerate}

The key technical result shown in this paper, needed to
obtain statistical guarantees for ridge regression, is the following property of
regularized volume sampling (where 
$d_\lambda$ is defined as in (\ref{eq:statistical-dimension})): 

\begin{theorem}\label{thm:square-inverse}
For any $\X\in\R^{d\times n}$, $\lambda\geq 0$ and $s\geq d_\lambda$,
let $S$ be sampled according  to
$\lambda$-regularized size $s$ volume sampling from $\X$. Then,
\[\E_S\, (\X_S\X_S^\top+\lambda\I)^{-1} \preceq
  \frac{n-d_\lambda+1}{s-d_\lambda+1}(\X\X^\top+\lambda\I)^{-1},\]
where $\preceq$ denotes a positive semi-definite inequality between
matrices. 
\end{theorem}

As a consequence of Theorem \ref{thm:square-inverse}, we show that
ridge estimators computed from volume sampled subproblems offer
statistical guarantees with respect to the full regression problem
$(\X,\y)$, despite observing only a small portion of the labels.

\begin{theorem}\label{thm:ridge}
Let $\X\in\R^{d\times n}$ and $\w^*\in\R^d$, and suppose that
$\y=\X^\top\w^* + \xib$, where $\xib$ is a mean zero vector with
$\Var[\xib]\preceq\sigma^2\,\I$. Let $S$ be sampled according to
$\lambda$-regularized size $s\geq d_\lambda$ volume sampling from $\X$ and
$\of{\wbh_\lambda^*}S$ be 
the $\lambda$-ridge estimator of $\w^*$ computed from subproblem
$(\X_S,\y_S)$. Then, if $\lambda\leq \frac{\sigma^2}{\|\w^*\|^2}$, we have
\begin{align*}
&\text{\bf(MSPE)}\quad   \E_S\E_{\xib}\,\frac{1}{n}\|\X^\top(\of{\wbh_\lambda^*}S-\w^*)\|^2\leq\frac{\sigma^2 d_\lambda}{s-d_\lambda+1},\\
&\text{\bf(MSE)}\
\E_S\E_{\xib}\|\of{\wbh_\lambda^*}S-\w^*\|^2 \!\leq
  \frac{\sigma^2n\,\tr((\X\X^\top\!\!+\!\lambda\I)^{-1})}{s-d_\lambda+1}.
\end{align*}
\end{theorem}

Next, we present two lower-bounds for MSPE of a
subsampled ridge estimator which show that the statistical guaranties
achieved by regularized volume sampling are
nearly optimal for $s\gg d_\lambda$ and better than standard
approaches for $s=O(d_\lambda)$. In particular, we show that 
non-i.i.d.~nature of volume sampling is essential
if we want to achieve good generalization when the number of
labels is close to $d_\lambda$. 
Namely, for certain data matrices, any subsampling procedure
selecting examples in an i.i.d.~fashion (e.g., leverage score sampling), requires more than $d_\lambda\ln(d_\lambda)$ labels to
achieve MSPE below $\sigma^2$, whereas volume
sampling obtains that bound for any matrix with $2d_\lambda$ labels.

\begin{theorem}\label{thm:lb-all}
For any $p\geq 1$ and $\sigma\geq 0$, there is $d\geq p$ such
that for any sufficiently large $n$ divisible by $d$ there exists a
matrix $\X\in\R^{d\times n}$ such that 
\[d_\lambda(\X)\geq p\quad \text{ for any }\quad 0\leq\lambda\leq \sigma^2,\] 
and for each of the following two statements there is a vector
$\w^*\in\R^d$ for which the corresponding
 regression problem $\y=\X^\top\w^*+\xib$ with $\Var[\xib]=\sigma^2\I$
satisfies that statement:
\vspace{-1mm}
\begin{enumerate}
\item For any subset $S\subseteq\{1..n\}$ of size $s$,
\begin{align*}
  \E_{\xib}\,\frac{1}{n}\|\X^\top(\of{\wbh_\lambda^*}S-\w^*)\|^2 
&\geq\frac{\sigma^2 d_\lambda}{s+d_\lambda};
\end{align*}
\item For multiset $S\subseteq\{1..n\}$ of size $s\leq
  d_\lambda(\ln(d_\lambda)-1)$, sampled i.i.d. from any distribution
  over $\{1..n\}$,
\begin{align*}
\E_S\E_{\xib}\, \frac{1}{n}\|\X^\top(\of{\wbh_\lambda^*}S-\w^*)\|^2 \geq \sigma^2.
\end{align*}
\end{enumerate}
\end{theorem}

% \begin{theorem}
% For any $n,d,\sigma$ and $\lambda\geq 0$, there is a matrix $\X\in\R^{d\times
%   n}$ and vector $\w^*\in\R^d$ such that for a multiset
% $S\subseteq\{1..n\}$ of $s\leq d_\lambda\log(d_\lambda)$ points
% sampled i.i.d. from any distribution, with probability at least 1/2 we
% have 
% \begin{align*}
% \E_{\xib}\frac{1}{n}\|\X^\top(\of{\wbh_\lambda^*}S-\w^*)\|^2& \geq \sigma^2.
% % \text{and}\quad
% % \E_{\xib}\frac{1}{n}\|\of{\wbh_\lambda^*}S-\w^*\|^2 &\geq \sigma^2
% % \tr((\X\X^\top+\lambda\I)^{-1})
% \end{align*}
% \end{theorem}

Finally, we propose an algorithm for regularized volume
sampling which runs in time $O((n+d)d^2)$. For the previously studied
case of $\lambda=0$, this algorithm offers a significant asymptotic
speed-up over existing volume sampling
algorithms (both exact and approximate).  

\begin{theorem}
\label{thm:exact}
For any $\lambda,\delta,s\geq 0$, there is an algorithm sampling according to
$\lambda$-regularized size $s$ volume sampling, that with probability at least
$1-\delta$ runs in time\footnote{We are primarily interested in the
  case where $n\geq d$. However, if $n<d$ and $\lambda>0$, then our
  techniques can be adapted to obtain an $O(n^2d)$ algorithm.}
$$O\!\left(\left(n+d+\log\left(\frac{n}{d}\right)
    \log\left(\frac{1}{\delta}\right)\right)d^2\right).$$ 
\end{theorem}
% The proof of Theorem \ref{thm:exact} is based on a new implementation of
% Reverse Iterative Volume Sampling, a procedure introduced by
% \cite{unbiased-estimates}.
When $n>d$ the time complexity of our proposed
algoirthm is not deterministic, but its dependence on the failure
probability $\delta$ is very small -- 
even for $\delta=2^{-n}$ the time complexity is still
$\Ot(nd^2)$. 

The remainder of this paper is arranged as follows: in Section
\ref{sec:statistical-guarantees} we present statistical analysis
of regularized volume sampling in the context of ridge regression, in
particular proving Theorems \ref{thm:square-inverse},
\ref{thm:ridge} and \ref{thm:lb-all}; in Section \ref{sec:algorithms} we present two
algorithms for regularized volume sampling and use them to prove
Theorem \ref{thm:exact}; in Section \ref{sec:experiments} we evaluate
the runtime of our algorithms on several standard linear regression datasets, and
compare the prediction performance of the subsampled ridge estimator under
volume sampling versus leverage score sampling; in Section
\ref{sec:conclusions} we summarize  the results and suggest 
future research directions.

\section{Statistical guarantees}
\label{sec:statistical-guarantees}

In this section, we show upper and lower bounds for the generalization
performance of subsampled ridge estimators, starting with an
important property of regularized volume sampling which connects it
with ridge regression. We will use
$\of{\Z_\lambda}S=\X_S\X_S^\top+\lambda\I$ as a shorthand in the
proofs.

\subsection{Proof of Theorem \ref{thm:square-inverse}}
To obtain this result, we will show how the expectation of matrix
$(\X_S\X_S^\top+\lambda\I)^{-1}$ changes when iteratively
removing a column in $\lambda$-Regularized Volume Sampling (see (\ref{eq:regularized})):
\begin{lemma}
Let $\X\in\R^{d\times n}$ and $S\subseteq\{1..n\}$.\\[2mm]
% Define probability $P(i|S) =
% \frac{\det(\X_\Sm\X_\Sm^\top+\lambda\I)}{\det(\X_S\X_S^\top+\lambda\I)}$
% for $i\in S$. Then,
If we sample $i\in S$ w.p. $\propto$
$\frac{\det(\X_\Sm\X_\Sm^\top+\lambda\I)}{\det(\X_S\X_S^\top+\lambda\I)}$,
then\\[-2mm]
\begin{align*}
\E_i\,(\X_\Sm\X_\Sm^\top+\lambda\I)^{-1}\!\preceq 
  \frac{s-d_\lambda+1}{s-d_\lambda} (\X_S\X_S^\top+\lambda\I)^{-1}.
\end{align*}
\end{lemma}
\proof 
We write the unnormalized probability of $i$ as:
\begin{align*}
h_i(S)&=\frac{\det(\of{\Z_\lambda}\Sm)}
        {\det(\of{\Z_\lambda}S)} \overset{(*)}{=} 1-\x_i^\top\of{\Z_\lambda}S^{-1}\x_i,
\end{align*}
where $(*)$ follows from Sylvester's theorem.
Next, letting $M=\sum_{i\in S}h_i(S)$, we compute
unnormalized expectation by applying the Sherman-Morrison formula to
$\of{\Z_\lambda}\Sm^{-1}$: 
\begin{align*}
M&\, \E_i\,(\X_\Sm\X_\Sm^\top+\lambda\I)^{-1}=\sum_{i\in S} h_i(S)
                                                         \of{\Z_\lambda}\Sm^{-1}
  \\ 
&= \sum_{i\in S} h_i(S)\left(\of{\Z_\lambda}S^{-1} +
                                         \frac{\of{\Z_\lambda}S^{-1}\x_i\x_i^\top
                                         \of{\Z_\lambda}S^{-1}}{1-\x_i^\top\of{\Z_\lambda}S^{-1}\x_i}\right)\\
&=M\, \of{\Z_\lambda}S^{-1} + \of{\Z_\lambda}S^{-1}\left(\sum_{i\in S}\x_i\x_i^\top\right) \of{\Z_\lambda}S^{-1}\\
&=M\, \of{\Z_\lambda}S^{-1} + \of{\Z_\lambda}S^{-1}(\of{\Z_\lambda}S -
  \lambda\I)\of{\Z_\lambda}S^{-1}\\
&=M\, \of{\Z_\lambda}S^{-1} + \of{\Z_\lambda}S^{-1} -
  \lambda\of{\Z_\lambda}S^{-2}\\
&\preceq (M+1)\, \of{\Z_\lambda}S^{-1}.
\end{align*}
Finally, we compute the normalization factor:
\begin{align*}
M &= \sum_{i\in S}(1-\x_i^\top\of{\Z_\lambda}S^{-1}\x_i) \\
&= s -
    \tr(\X_S^\top\of{\Z_\lambda}S^{-1}\X_S) =s -
  \tr(\X_S\X_S^\top\of{\Z_\lambda}S^{-1}) \\
&=s-\tr((\of{\Z_\lambda}S-\lambda\I)\of{\Z_\lambda}S^{-1}) \\
&=s - \tr(\I) + \lambda\tr(\of{\Z_\lambda}S^{-1})\\
&\geq s-d + \lambda\tr(\of{\Z_\lambda}{\{1..n\}}^{-1}) = s-d_\lambda,
\end{align*}
where we used the fact that $d_\lambda$ can we rewritten as
$d - \lambda\tr(\of{\Z_\lambda}{\{1..n\}}^{-1}).$
Putting the bounds together, we obtain the result.
\hfill\BlackBox

To prove Theorem \ref{thm:square-inverse} it remains to chain the
conditional expectations along the sequence of subsets obtained by
$\lambda$-Regularized Volume Sampling:
\begin{align*}
\E_S\,\of{\Z_\lambda}S^{-1} &\preceq
  \left(\prod_{t=s+1}^n\frac{t-d_\lambda+1}{t-d_\lambda}\right)\,\of{\Z_\lambda}{\{1..n\}}^{-1}\\
&=\frac{n-d_\lambda+1}{s-d_\lambda+1}(\X\X^\top+\lambda\I)^{-1}.
\end{align*}

\subsection{Proof of Theorem \ref{thm:ridge}}
Standard analysis for the ridge regression estimator follows by
performing bias-variance decomposition of the error, and then
selecting $\lambda$ so that bias can be appropriately bounded. We
will recall this calculation for a fixed subproblem
$(\X_S,\y_S)$. First, we
compute the bias of the ridge estimator for a fixed set $S$:
\begin{align*}
\text{Bias}_{\xib}[\of{\wbh_\lambda^*}S] &= \E[\of{\wbh_\lambda^*}S] - \w^*\\
&=\E_{\xib}\,[\of{\Z_\lambda}S^{-1}\X_S\y_S] - \w^* \\
&=\of{\Z_\lambda}S^{-1}\X_S\,(\X_S^\top\w^* + \cancel{\E_{\xib}[\xib_S]}) - \w^*\\
&=(\of{\Z_\lambda}S^{-1}\X_S\X_S^\top - \I)\w^*\\
%[-4mm]
% &= \of{\Z_\lambda}S^{-1}\X_S\X_S^\top \
%   \E_{\xib}[\overbrace{(\X_S\X_S^\top)^{-1}\X_S\y_S}^{\wbh_0^*\!(S)}]\\ 
%&= \of{\Z_\lambda}S^{-1}(\of{\Z_\lambda}S - \lambda\I)\w^* - \w^* \\
&= -\lambda \of{\Z_\lambda}S^{-1}\w^*.
\end{align*}
Similarly, the covariance matrix of $\of{\wbh_\lambda^*}S$ is given by:
\begin{align*}
\Var_{\xib}[\of{\wbh_\lambda^*}S] &=
  \of{\Z_\lambda}S^{-1}\X_S\Var_{\xib}[\xib_S]\X_S^\top
  \of{\Z_\lambda}S^{-1}\\
&\preceq \sigma^2\of{\Z_\lambda}S^{-1}\X_S\X_S^\top
  \of{\Z_\lambda}S^{-1}\\
&=\sigma^2(\of{\Z_\lambda}S^{-1}-\lambda
  \of{\Z_\lambda}S^{-2}).
\end{align*}
Mean squared error of the ridge estimator for a fixed subset $S$ can
now be bounded by:
\begin{align}
\E_{\xib}\|&\of{\wbh_\lambda^*}S - \w^*\|^2 =
  \tr(\Var_{\xib}[\of{\wbh_\lambda^*}S]) + \|\text{Bias}_{\xib}[\of{\wbh_\lambda^*}S]\|^2\nonumber\\ %\|\E_{\xib}[\of{\wbh_\lambda^*}S]  - \w^*\|^2\\
\leq&\sigma^2\tr(\of{\Z_\lambda}S^{-1}\!\!\!-\lambda
  \of{\Z_\lambda}S^{-2}) +
  \lambda^2\tr(\of{\Z_\lambda}S^{-2}\w^*\w^{*\top})\nonumber\\
\leq& \sigma^2\tr(\of{\Z_\lambda}S^{-1}) +
  \lambda\tr(\of{\Z_\lambda}S^{-2})(\lambda\|\w^*\|^2\!- \sigma^2)\label{eq:cauchy-trace}\\
\leq& \sigma^2\tr(\of{\Z_\lambda}S^{-1}),\label{eq:lambda-bound}
\end{align}
where in (\ref{eq:cauchy-trace}) we applied Cauchy-Schwartz inequality for matrix trace, and
in (\ref{eq:lambda-bound}) we used the assumption that $\lambda\leq
\frac{\sigma^2}{\|\w^*\|^2}$. Thus, taking expectation over the sampling of set $S$, we get
\begin{align}
\E_S\E_{\xib}&\|\of{\wbh_\lambda^*}S - \w^*\|^2 \leq
  \sigma^2\E_S\tr(\of{\Z_\lambda}S^{-1}) \nonumber\\
\text{(Thm. $1$)}\ &\leq\sigma^2
  \frac{n-d_\lambda+1}{s-d_\lambda+1}
  \tr(\of{\Z_\lambda}{\{1..n\}}^{-1})\label{eq:ridge-mse}\\
&\leq \frac{\sigma^2\,n\, \tr((\X\X^\top+\lambda\I)^{-1})}{s-d_\lambda+1}.\nonumber
\end{align}
% Note that to bound the mean squared error, we only needed to bound the
% expectation of the trace of matrix $\of{\Z_\lambda}S^{-1}$. Theorem
% \ref{thm:square-inverse} offers a much more general bound using the
% positive semi-definite inequality. To illustrate the strength of that statement,
% we now show a bound on MSPE of the ridge
% estimator, which is also a straight-forward consequence of Theorem
% \ref{thm:square-inverse}, but it does not follow from its trace version.
Next, we bound the mean squared prediction error. As before, we start
with the standard  bias-variance decomposition for fixed set $S$:
\begin{align*}
\E_{\xib}&\|\X^\top(\of{\wbh_\lambda^*}S - \w^*)\|^2 \\
&=
  \tr(\Var_{\xib}[\X^\top\of{\wbh_\lambda^*}S]) + \|\X^\top(\E_{\xib}[\of{\wbh_\lambda^*}S]
  - \w^*)\|^2\\
&\leq\sigma^2\tr(\X^\top(\of{\Z_\lambda}S^{-1}\!\!\!-\lambda
  \of{\Z_\lambda}S^{-2})\X) \\
&\quad +
  \lambda^2\tr(\of{\Z_\lambda}S^{-1}\X\X^\top \of{\Z_\lambda}S^{-1}\w^*\w^{*\top})\\
&\leq \sigma^2\tr(\X^\top\of{\Z_\lambda}S^{-1}\X)\\
&\quad + \lambda\tr(\X^\top\of{\Z_\lambda}S^{-2}\X)(\lambda\|\w^*\|^2
  - \sigma^2)\\
&\leq \sigma^2\tr(\X^\top\of{\Z_\lambda}S^{-1}\X).
\end{align*}
Once again, taking expectation over subset $S$, we have
\begin{align}
\E_S&\E_{\xib}\frac{1}{n}\|\X^\top(\of{\wbh_\lambda^*}S -
  \w^*)\|^2\nonumber\\
&\leq\frac{\sigma^2}{n}
  \E_S\,\tr(\X^\top\of{\Z_\lambda}S^{-1}\X)\nonumber\\
&=\frac{\sigma^2}{n}\tr(\X^\top \E_S[\of{\Z_\lambda}S^{-1}]\X)\nonumber\\
\text{(Thm. $1$)}\ &\leq \frac{\sigma^2}{n} \frac{n-d_\lambda+1}{s-d_\lambda+1}\tr(\X^\top
  \of{\Z_\lambda}{\{1..n\}}^{-1}\X)\label{eq:ridge-mspe}\\
& \leq \frac{\sigma^2 d_\lambda}{s-d_\lambda+1}.\nonumber
\end{align}
The key part of proving both bounds is the application of Theorem
\ref{thm:square-inverse}. For MSE, we only used the trace version of the
inequality (see (\ref{eq:ridge-mse})), however to obtain the bound on
MSPE we used the more general positive semi-definite inequality in
(\ref{eq:ridge-mspe}).

\subsection{Proof of Theorem \ref{thm:lb-all}}
Let $d=\lceil p\rceil+1$ and $n\geq\lceil \sigma^2\rceil d(d-1)$ be
divisible by $d$. We define 
\begin{align*}
\X &\defeq [\I,...,\I]\in\R^{d\times n}\\
\w^{*\top}&\defeq \,[a\sigma,...,a\sigma]\in\R^d
\end{align*}
 for some $a>0$. For any  $\lambda\leq \sigma^2$, the
 $\lambda$-statistical dimension of $\X$ is 
\begin{align*}
d_\lambda &= \tr(\X^\top\of{\Z_\lambda}{\{1..n\}}^{-1}\X)\\
&\geq\frac{\lceil \sigma^2\rceil d(d-1)}{\lceil \sigma^2\rceil (d-1) +
            \lambda}\geq \frac{d(d-1)}{d-1+1}\geq p.
\end{align*}

Let $S\subseteq\{1..n\}$ be any set of size $s$, and for $i\in\{1..d\}$ let 
\[s_i \defeq |\{i\in S:\, \x_i=\e_i\}|. \]
The prediction variance of estimator $\of{\wbh_\lambda^*}S$ is equal to
\begin{align*}
\tr(\Var_{\xib}[\X^\top\of{\wbh_\lambda^*}S])
&=\sigma^2\tr(\X^\top\!(\of{\Z_\lambda}S^{-1}\!\!-\lambda \of{\Z_\lambda}S^{-2})\X)\\
&=\frac{\sigma^2n}{d}\sum_{i=1}^d\left(\frac{1}{s_i+\lambda} -
  \frac{\lambda}{(s_i+\lambda)^2}\right)\\
&=\frac{\sigma^2n}{d}\sum_{i=1}^d\frac{s_i}{(s_i+\lambda)^2}.
% &\geq\frac{n}{\frac{s}{d}+\lambda} - \frac{n}{(\frac{s}{d}+\lambda)^2}\\
% & = \frac{n\,d}{s + \lambda d}\left(1 - \frac{\lambda d}{s+\lambda
%   d}\right)\\
% &=\frac{n\,d}{s}\left(1 - \frac{\lambda d}{s+\lambda d}\right)^2
\end{align*}

The prediction bias of estimator $\of{\wbh_\lambda^*}S$ is equal to
\begin{align*}
\|\X^\top&(\E_{\xib}[\of{\wbh_\lambda^*}S]-\w^*)\|^2\\
&=\lambda^2\w^{*\top}\of{\Z_\lambda}S^{-1}\X\X^\top\of{\Z_\lambda}S^{-1}\w^{*\top}\\
&=\frac{\lambda^2a^2\sigma^2n}{d}\,\tr(\of{\Z_\lambda}S^{-2})\\
&=\frac{\lambda^2a^2\sigma^2n}{d}\sum_{i=1}^d\frac{1}{(s_i+\lambda)^2}.
\end{align*}

Thus, MSPE of estimator  $\of{\wbh_\lambda^*}S$
is equal to:
\begin{align*}
\E_{\xib}&\frac{1}{n}\|\X^\top(\of{\wbh_\lambda^*}S-\w^*)\|^2\\
&=\frac{1}{n}\tr(\Var_{\xib}[\X^\top\of{\wbh_\lambda^*}S])
  +\frac{1}{n}\|\X^\top(\E_{\xib}[\of{\wbh_\lambda^*}S]-\w^*)\|^2\\
&=\frac{\sigma^2}{d}\sum_{i=1}^d\left(\frac{s_i}{(s_i+\lambda)^2} +
  \frac{a^2\lambda^2}{(s_i+\lambda)^2}\right)\\
&=\frac{\sigma^2}{d}\sum_{i=1}^d\frac{s_i +a^2\lambda^2}{(s_i+\lambda)^2}.
% &=\frac{\sigma^2}{n}\tr(\X^\top(\of{\Z_\lambda}S^{-1}-\lambda\of{\Z_\lambda}S^{-2})\X)\\
% &=\frac{\sigma^2}{n}\left(\frac{n}{\frac{s}{d}+\lambda} -
%   \frac{\lambda n}{(\frac{s}{d}+\lambda)^2}\right)\\
% &\geq\frac{\sigma^2d}{s}\left(1-\frac{\lambda d}{s+\lambda d}\right)^2\\
% &\geq\frac{\sigma^2d_\lambda}{s}\left(1-\frac{\lambda
% (d_\lambda+1)}{s+\lambda 
% d_\lambda}\right)^2
%   \frac{\sigma^2d_\lambda}{s-d_\lambda+1}\left(1-\frac{\lambda}{s+\lambda}\right)
%   \left(1-\frac{1}{s+\lambda}\right)\\
% &\geq \frac{\sigma^2d_\lambda}{s-d_\lambda+1}\left(1 - \frac{1+\lambda}{s+\lambda}\right).
\end{align*}

Next, we find the $\lambda$ that minimizes this expression. Taking the
derivative with respect to $\lambda$ we get:
\begin{align*}
\frac{\partial}{\partial
  \lambda}\left(\frac{\sigma^2}{d}\sum_{i=1}^d\frac{s_i
  +a^2\lambda^2}{(s_i+\lambda)^2}\right)=\frac{\sigma^2}{d}\sum_{i=1}^d\frac{2s_i(\lambda
  - a^{-2})}{(s_i+\lambda)^3}.
\end{align*}

Thus, since at least one $s_i$ has to be greater than $0$, for any set
$S$ the derivative is negative  for $\lambda< a^{-2}$ and
positive for $\lambda>a^{-2}$, and the unique minimum of
MSPE is achieved at $\lambda=a^{-2}$, regardless of 
which subset $S$ is chosen. So, as we are seeking a lower bound, we can
focus on the case of $\lambda=a^{-2}$. 

\subsubsection{Proof of Part 1}
Let $a=1$. As shown above, we can assume that $\lambda=1$. In this
case the formula simplifies to:
\begin{align*}
\E_{\xib}&\frac{1}{n}\|\X^\top(\of{\wbh_\lambda^*}S-\w^*)\|^2 =\frac{\sigma^2}{d}\sum_{i=1}^d\frac{s_i+1}
    {(s_i+1)^2}\\ 
&=\frac{\sigma^2}{d}\sum_{i=1}^d\frac{1}{s_i+1}\overset{(*)}{\geq}
  \frac{\sigma^2}{\frac{s}{d} +
  1}=\frac{\sigma^2d}{s+d}\geq \frac{\sigma^2d_\lambda}{s+d_\lambda},
\end{align*}
where $(*)$ follows by applying Jensen's inequality to convex function
$\phi(x)=\frac{1}{x+1}$. 

\subsubsection{Proof of Part 2}
Let $a=\sqrt{2d}$. As shown above, we can assume that
$\lambda=1/(2d)$. Suppose that set $S$ is sampled i.i.d. from some
distribution over set $\{1..n\}$. Using standard analysis for the
Coupon Collector's problem, it can be shown that if $|S|\leq
d(\ln(d)-1)$, then with probability at least
$1/2$ there is $i\in\{1..d\}$ such that $s_i=0$ (ie, one of
the unit vectors $\e_i$ was never selected). Thus,
MSPE can be lower-bounded as follows:
\begin{align*}
\E_S\E_{\xib}&\frac{1}{n}\|\X^\top(\of{\wbh_\lambda^*}S-\w^*)\|^2\\
&\geq
  \frac{1}{2}\,\frac{\sigma^2}{d}\,
  \frac{s_i+a^2\lambda^2}{(s_i+\lambda)^2}=\frac{\sigma^2}{2d}
  \frac{2d \lambda^2}{\lambda^2} = \sigma^2.
\end{align*}

\section{Algorithms}
\label{sec:algorithms}
In this section, we present two algorithms that implement
$\lambda$-regularized volume sampling \eqref{eq:regularized}. 
The first one, RegVol (Algorithm
\ref{alg:volsamp}), simply adds regularization to the
algorithm proposed in \cite{unbiased-estimates}
for Reverse Iterative Volume Sampling \eqref{eq:reviter}. 
Note that in this algorithm the conditional
distribution $P(i\,|\,S)$ is updated at every iteration, which leads
to $O((n^2\!+\!d^2)d)$ time complexity. The second
algorithm, FastRegVol (Algorithm \ref{alg:lazyvolsamp}), is
our new algorithm which avoids recomputing the conditional distribution
at every step, making it essentially as fast as exact leverage score sampling.
% Note, that the conditional distribution $P(i\,|\,S)$ changes after each
% iteration of $\lambda$-Regularized Volume Sampling. Updating it is the
% dominant computational 
% cost of the algorithm proposed by \cite{unbiased-estimates} for
% volume sampling with $\lambda=0$, running in
% time $O(n^2d)$. This approach can be easily adapted to sample
% according to any $\lambda$ as seen in Algorithm
% \ref{alg:volsamp}, called {\volsamp}.
\begin{algorithm}
  \caption{{\volsamp}($\X,s,\lambda$) (adapted from \cite{unbiased-estimates})}
  \begin{algorithmic}[1]
%    \STATE {\volsamp}($\X,s$):
    \STATE $\Z\leftarrow (\X\X^\top+\lambda\I)^{-1}$\label{line:inv}   % - $O(nd^2)$ 
%    \FOR {$j=1..n$}
    \STATE $\forall_{i\in\{1..n\}} \quad h_i\leftarrow 1-\x_i^\top \Z\x_i$
%    \quad- $O(d^2)$
%   \ENDFOR
    \STATE $S \leftarrow \{1..n\}$
    \STATE {\bf while} $|S|>s$
    \STATE \quad Sample $i \propto h_i$ out of $S$
    \STATE \quad $S\leftarrow S - \{i\}$
    \STATE \quad $\v \leftarrow \Z\x_i /\sqrt{h_i}$
    \STATE \quad $\forall_{j\in S}\quad  h_j\leftarrow h_j -  (\x_j^\top\v)^2$
    % \ENDFOR
    \STATE \quad $\Z \leftarrow \Z + \v\v^\top$  %  - $O(d^2)$
    \STATE {\bf end} 
    \RETURN $S$
 \end{algorithmic}
\label{alg:volsamp}
\end{algorithm}

\begin{algorithm}%[H] 
% {\fontsize{8}{8}\selectfont
 \caption{{\lazy}($\X,s,\lambda$)}
\label{alg:lazyvolsamp}
\begin{algorithmic}[1]
%  \STATE \textbf{Input:} $\X\!\in\!\R^{d\times n}$, $s\!\in\!\{d..n\}$ 
%  \STATE $\Z\leftarrow (\X\X^\top)^{-1}$\label{line:inv}   % - $O(nd^2)$ 
  % \FOR {$j=1..n$}
%  \STATE $\forall_{i\in\{1..n\}} \quad h_i\leftarrow 1-\x_i^\top \Z\x_i$
  % \quad- $O(d^2)$
  % \ENDFOR
  %\STATE $S \leftarrow \{1,..,n\}$
%  \STATE AdaVol[adaptive]($\X,s$):
  \STATE $\Z\leftarrow (\X\X^\top+\lambda\I)^{-1}$\label{line:inv}   % - $O(nd^2)$ 
%    \FOR {$j=1..n$}
%\STATE $\forall_{i\in\{1..n\}} \quad \hat{h}_i\leftarrow 1$
%  \Z\x_i$
  \STATE $S \leftarrow \{1..n\}$
  \STATE {\bf while} $|S|>s$
  \STATE \quad \textbf{repeat}
%  \STATE \quad\quad Sample $i \propto \hat{h}_i$ out of $S$
  \STATE \quad\quad Sample $i$ uniformly out of $S$
  \STATE \quad\quad $h_i \leftarrow 1-\x_i^\top \Z\x_i$
%  \STATE \quad\quad Sample $A \sim \text{Bernoulli}(h_i/\hat{h}_i)$
\STATE \quad\quad Sample $A \sim \text{Bernoulli}(h_i)$
%  \STATE \quad\quad $\hat{h}_i \leftarrow h_i$
  \STATE \quad \textbf{until} $A=1$
  \STATE \quad $S\leftarrow S - \{i\}$
%  \STATE \quad $\v \leftarrow \Z\x_i /\sqrt{h_i}$
%  \STATE \quad $\Z \leftarrow \Z + \v\v^\top$  %  - $O(d^2)$
  \STATE \quad $\Z \leftarrow \Z + h_i^{-1}\Z\x_i\x_i^\top\Z$  %  - $O(d^2)$
  \STATE {\bf end} 
  \RETURN $S$
\end{algorithmic}
\end{algorithm}

 The correctness of RegVol follows from Lemma
\ref{lem:weight-bound}, which is a straight-forward application of the
Sherman-Morrison formula (see \cite{unbiased-estimates} for more details).

\begin{lemma}\label{lem:weight-bound}
For any matrix $\X\in \R^{d\times n}$, set $S\subseteq\{1..n\}$ and two distinct
indices $i,j\in S$, we have
\begin{align*}
1-\x_j^\top\of{\Z_\lambda}{\Sm}^{-1}\x_j 
%= 1- \x_j^\top\XinvS\x_j -  (\x_j^\top\XinvS\x_i)^2
= 1- \x_j^\top \of{\Z_\lambda}S^{-1}\x_j - a^2,
\end{align*}
where 
\[a = \frac{\x_j^\top \of{\Z_\lambda}S^{-1}\x_i}{\sqrt{1-\x_i^\top \of{\Z_\lambda}S^{-1}\x_i}}.\]
\end{lemma}
We propose a new volume sampling algorithm, which runs in time
$O((n\!+\!d)d^2)$, significantly faster than {\volsamp}
when $n\gg d$. Our key observation is that updating the full
conditional distribution $P(i|S)$ is wasteful, since the distribution
changes very slowly throughout the procedure. Moreover, the
unnormalized weights $h_i$, which are computed in the process are all
bounded by 1.
% Moreover, as seen in
% Lemma \ref{lem:weight-bound}, the
% unnormalized weights, which are computed in
% the process, can only decrease as we proceed with the
% algorithm.
% The immediate consequence of this fact is
% that the initial set of weights computed for the full dataset forms
% an upperbound over all of the conditional distributions that will be
% needed over the course of the algorithm.
% \begin{corollary}
% For any matrix  $\X\in \R^{d\times n}$, set $S\subseteq\{1..n\}$ and
% $i\in S$ we have
% \begin{align*}
% 1-\x_i^\top \of{\Z_\lambda}S^{-1}\x_i\leq 1 - \x_i^\top \of{\Z_\lambda}{\{1..n\}}^{-1}\x_i.
% \end{align*}
% \end{corollary}
%  Suppose that at some iteration of the {\volsamp} algorithm, we only have
%  upper-bound estimates $\hat{h}_i$ of the exact weights $h_i$ used
%  for sampling from the set $S$, ie:
% \begin{align*}
% \hat{h}_i\geq h_i= 1 - \x_i^\top\Z\x_i,\quad \forall i\in S.
% \end{align*}
Thus, to sample from the correct distribution at any given iteration,
we can employ rejection sampling as follows: 
%\vspace{-3mm}
\begin{enumerate}
\setlength\itemsep{0mm}
\item Sample $i$ uniformly from set $S$,
\item Compute $h_i$,
\item Accept with probability $h_i$,
\item Otherwise, draw another sample.
\end{enumerate}
%\vspace{-3mm}
Note that this rejection sampling can be employed locally, within each
iteration of the algorithm. Thus, one rejection does not revert us
back to the beginning of the algorithm.
Moreover, if the probability of acceptance is high, then this strategy
requires computing only a small number of weights per iteration of the algorithm, as
opposed to updating all of them. This turns out to be the case,
as shown below. The full pseudo-code of the sampling procedure is
given in Algorithm \ref{alg:lazyvolsamp}, called {\lazy}.
 \vspace{2mm}

% Note that in line 9 of {\lazy}
% a form of adaptive rejection sampling is employed, by updating the
% upper-bound distribution at the sampled point, which can only
% improve the acceptance rate for the remainder of the
% procedure. However, our analysis also applies to {\lazy} when line 9 is omitted. For
% experimental comparison of {\lazy} with and without adaptive sampling, see Section
% \ref{sec:experiments}. 

\subsection{Proof of Theorem \ref{thm:exact}}

To simplify the analysis, we combine {\volsamp} and {\lazy}, by
running {\lazy} while subset $S$ has size at least $2d$, and then, if
a smaller subset is needed, switching to {\volsamp}:
% \begin{algorithm}
%  \caption{Fast{\volsamp}($\X,s,\lambda$)}
% \label{alg:fastvolsamp}
\vspace{2mm}

\begin{algorithmic}[1]
  \STATE $S \leftarrow \text{{\lazy}}(\X,\max\{s,2d\},\lambda)$
  \STATE \textbf{if} $s < 2d$
  \STATE \quad $S \leftarrow \text{{\volsamp}}(\X_S,s,\lambda)$ %(see \cite{unbiased-estimates})
  \STATE \textbf{end}
  \RETURN $S$
\end{algorithmic}
% \end{algorithm}

Following the analysis of \cite{unbiased-estimates}, time
complexity of the {\volsamp} portion of the procedure is
$O((2\,d)^2d)=O(d^3)$. Next, we analyze the
efficiency of rejection sampling in the {\lazy} portion. Let $R_t$
be a random variable corresponding to the number of trials needed in
the \textbf{repeat} loop from line 4 in {\lazy} at the point when $|S|=t$. 
Note that conditioning on the algorithm's history,
$R_t$ is distributed according to geometric
distribution $\text{Ge}(q_t)$ with success probability:
\begin{align*}
q_t = \frac{1}{t}\sum_{i\in S}(1-\x_i^\top \of{\Z_\lambda}S^{-1}\x_i)\geq \frac{t-d}{t}\geq \frac{1}{2}.
\end{align*}
Thus, even though variables $R_t$ are not themselves independent, they
can be upper-bounded by a sequence of independent variables
$\Rh_t\sim \text{Ge}(\frac{t-d}{t})$. The expectation of the total
number of trials in {\lazy}, $\bar{R}=\sum_t R_t$, can thus be bounded
as follows: 
\begin{align*}
\E[\bar{R}]&\leq \sum_{t=2d}^n\E[\Rh_t]
             =\sum_{t=2d}^n\frac{t}{t-d}\leq 2n.
\end{align*}

Next, we will obtain a similar bound with
high probability instead of in expectation. Here, we will have to use
the fact that the variables $\Rh_t$ are independent, which means that
we can upper-bound their sum with high probability using standard
concentration bounds for geometric distribution. For example, using
Corollary 2.2 from 
\cite{geometric-tail-bounds} one can immediately show that with
probability at least $1-\delta$ we have $\bar{R} =
O(n\ln\delta^{-1})$. However, more careful analysis shows an even
better dependence on $\delta$. 

\begin{lemma}
\label{lm:tail-bound}
Let $\Rh_t\sim \textnormal{Ge}(\frac{t-d}{t})$ be independent random
variables. Then w.p. at least  $1-\delta$
\[\sum_{t=2d}^n \Rh_t = O\left(n +
    \log\left(\frac{n}{d}\right)\log\left(\frac{1}{\delta}\right)\right). \] 
\end{lemma}
\proof%\begin{proof}
As observed by \cite{geometric-tail-bounds}, tail-bounds for the sum
of geometric random variables depend on the minimum acceptance
probability among those variables. Note that for the vast majority of
$\Rh_t$'s the acceptance probability is very close to 1, so
intuitively we should be able to take advantage of this to improve our
tail bounds. To that end, we partition the variables into groups of
roughly similar acceptance probability and then separately bound the
sum of variables in each group. Let $J=\log(\frac{n}{d})$ (w.l.o.g. assume
that $J$ is an integer). For $1\leq j\leq J$, let
$I_j=\{d2^j, d2^j+1,..,d2^{j+1}\}$ represent the $j$-th
partition. We use the following notation for  each partition:
\begin{align*}
\bar{R}_j &\defeq \sum_{t\in I_j}R_t,
&  \mu_j &\defeq\E[\bar{R}_j],\\%\geq d2^{j}\\
r_j &\defeq \min_{t\in I_j}\frac{t-d}{t}, %\geq \frac{d2^j-d}{d2^j}= 1-\frac{1}{2^j} 
& \gamma_j &\defeq \frac{\log(\delta^{-1})}{d2^{j-2}} + 3.
\end{align*}
Now, we apply Theorem 2.3 of \cite{geometric-tail-bounds} to
$\bar{R}_j$, obtaining
\begin{align*}
P(\bar{R}_j&\geq \gamma_j\mu_j) \leq
  \gamma_j^{-1}(1-r_j)^{(\gamma_j-1-\ln\gamma_j)\mu_j} \\
&\overset{(1)}{\leq}  (1-r_j)^{\gamma_j\mu_j/4}  \overset{(2)}{\leq} 2^{-j\gamma_jd2^{j-2}},
\end{align*}
where $(1)$ follows since $\gamma_j\geq 3$, and $(2)$ holds because
$\mu_j\geq d2^j$ and $r_j\geq 1-2^{-j}$. Moreover, for the chosen
$\gamma_j$ we have
\begin{align*}
j\gamma_jd2^{j-2} &= j\log(\delta^{-1}) + 3jd2^{j-2} \\
&\geq \log(\delta^{-1}) + j = \log(2^j\delta^{-1}).
\end{align*}
Let $A$ denote the event that $\bar{R}_j\leq \gamma_j\mu_j$ for all
$j\leq J$. Applying union bound, we get
\begin{align*}
P(A) &\geq 1- \sum_{j=1}^J
  P(\bar{R}_j\geq \gamma_j\mu_j) \\
&\geq 1- \sum_{j=1}^J
       2^{-\log(2^j\delta^{-1})}=1-\sum_{j=1}^J \frac{\delta}{2^j}\geq
  1-\delta.
\end{align*}
If $A$ holds, then we obtain the desired bound:
\begin{align*}
\sum_{t=2d}^n \Rh_t&\leq \sum_{j=1}^J \gamma_j\mu_j \leq
  \sum_{j=1}^J\left(\frac{\log(\delta^{-1})}{d2^{j-2}} +
  3\right)\,d2^{j+1} \\
&=8 J \log(\delta^{-1}) + 6\sum_{j=1}^Jd2^j\\
&= O\left(\log\left(\frac{n}{d}\right)\log\left(\frac{1}{\delta}\right) +  n\right).
\hspace{2cm}\BlackBox
\end{align*}
%\end{proof}

Returning to the proof of Theorem \ref{thm:exact}, we note that each
trial of rejection sampling requires computing one weight $h_i$ in
time $O(d^2)$. The overall time complexity of FastVol also includes computation
and updating of matrix $\Z$ (in time $O((n+d)d^2)$), rejection sampling which
takes $O\left(\left(n+\log\left(\frac{n}{d}\right)\log\left(\frac{1}{\delta}\right)\right) d^2\right)$
time, and (if $s<2d$) the {\volsamp} portion, taking $O(d^3)$ (see
\cite{unbiased-estimates} for details). This concludes the proof of
Theorem \ref{thm:exact}.

\section{Experiments}
\label{sec:experiments} 
In this section we experimentally evaluate the proposed
algorithms for regularized volume sampling, in terms of runtime and
the quality of subsampled ridge estimators. 
 The list of
implemented algorithms is:
\begin{enumerate}
\item Regularized Volume Sampling:
\begin{enumerate}
\item {\lazy} -- Our new approach (see Alg. \ref{alg:lazyvolsamp});
\item {\volsamp} -- Adapted from \cite{unbiased-estimates} (see Alg. \ref{alg:volsamp});
\end{enumerate}
\item Leverage Score Sampling\footnote{Regularized variants of
    leverage scores have also been considered in context of kernel
    ridge regression \cite{ridge-leverage-scores}. However, in our experiments
    regularizing leverage scores did not provide any improvements.}
  (LSS) -- a popular i.i.d. sampling  technique
  \cite{randomized-matrix-algorithms}, where examples are selected
  w.p. $P(i) = (\x_i^\top(\X\X^\top)^{-1}\x_i)/d.$
\end{enumerate}
The experiments were performed on several benchmark linear regression 
datasets \cite{uci-repository}. Table
\ref{tab:datasets} lists those datasets along with running times
for sampling dimension many columns with each method.
Dataset MSD was too big for {\volsamp} to finish in reasonable time.
\begin{table}[H]
\begin{center}
\begin{tabular}{c|c|c|c|c}
Dataset & $d\times n$ & \!{\volsamp}\! &\!{\lazy}\! & LSS \\
\hline
cadata & $8\times 21$k  & 33.5s & 0.9s&0.1s\\
MSD & \!$90\!\times\!464$k\! & $>$24hr & 39s& 12s\\
\!cpusmall\! & $12\times 8$k &1.7s&0.4s&0.07s\\
abalone & $8\times 4k$ &0.5s&0.2s&0.03s
\end{tabular}
\caption{A list of used regression datasets, with runtime comparison
  between {\volsamp} (see Alg. \ref{alg:volsamp} and
  \cite{unbiased-estimates}) and {\lazy} (see
  Alg. \ref{alg:lazyvolsamp}). We also provide the 
  runtime for obtaining exact leverage score samples (LSS).}  
\label{tab:datasets}
\end{center}
\end{table}

\begin{figure}
\includegraphics[width=0.25\textwidth]{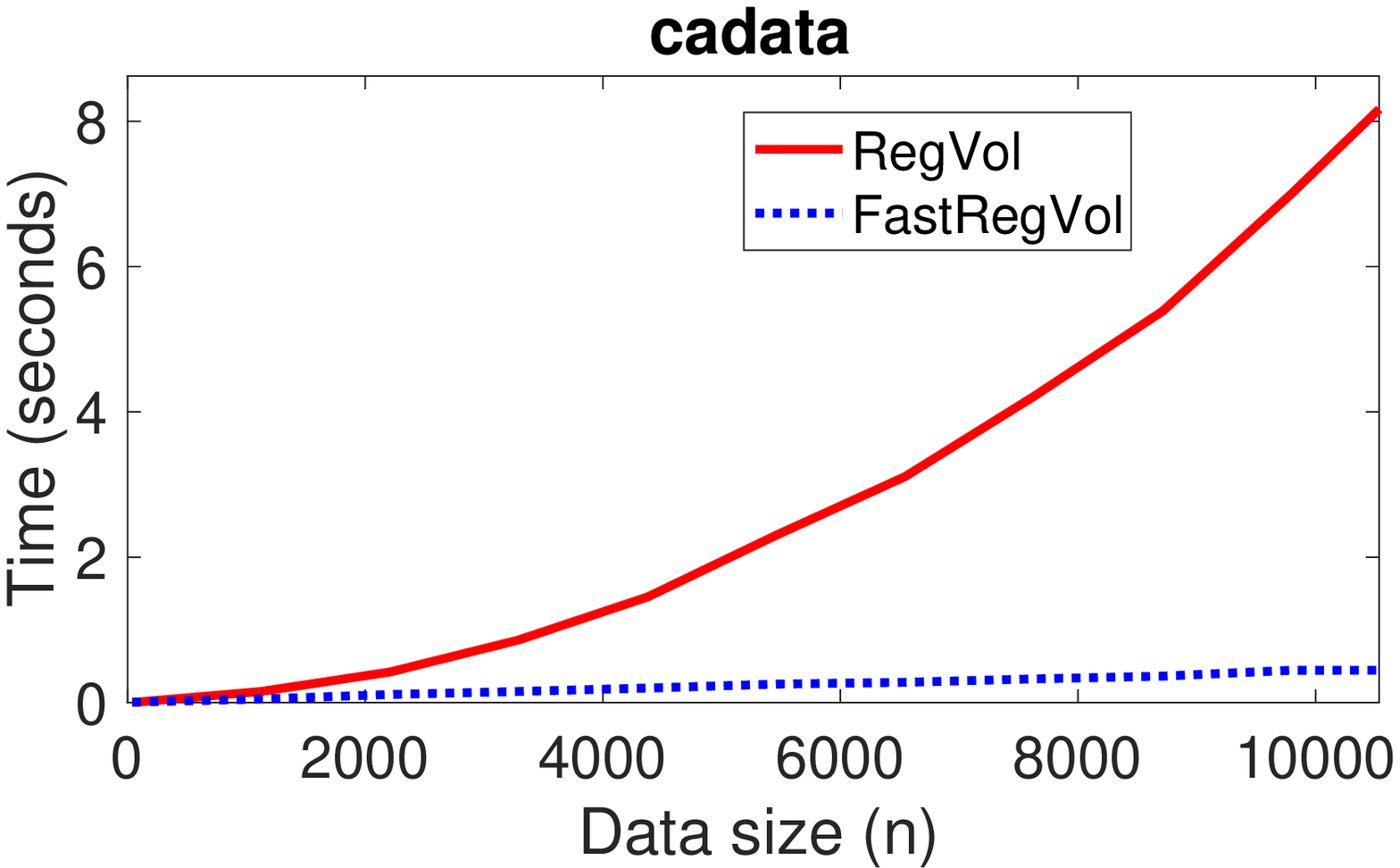}\nobreak
\includegraphics[width=0.25\textwidth]{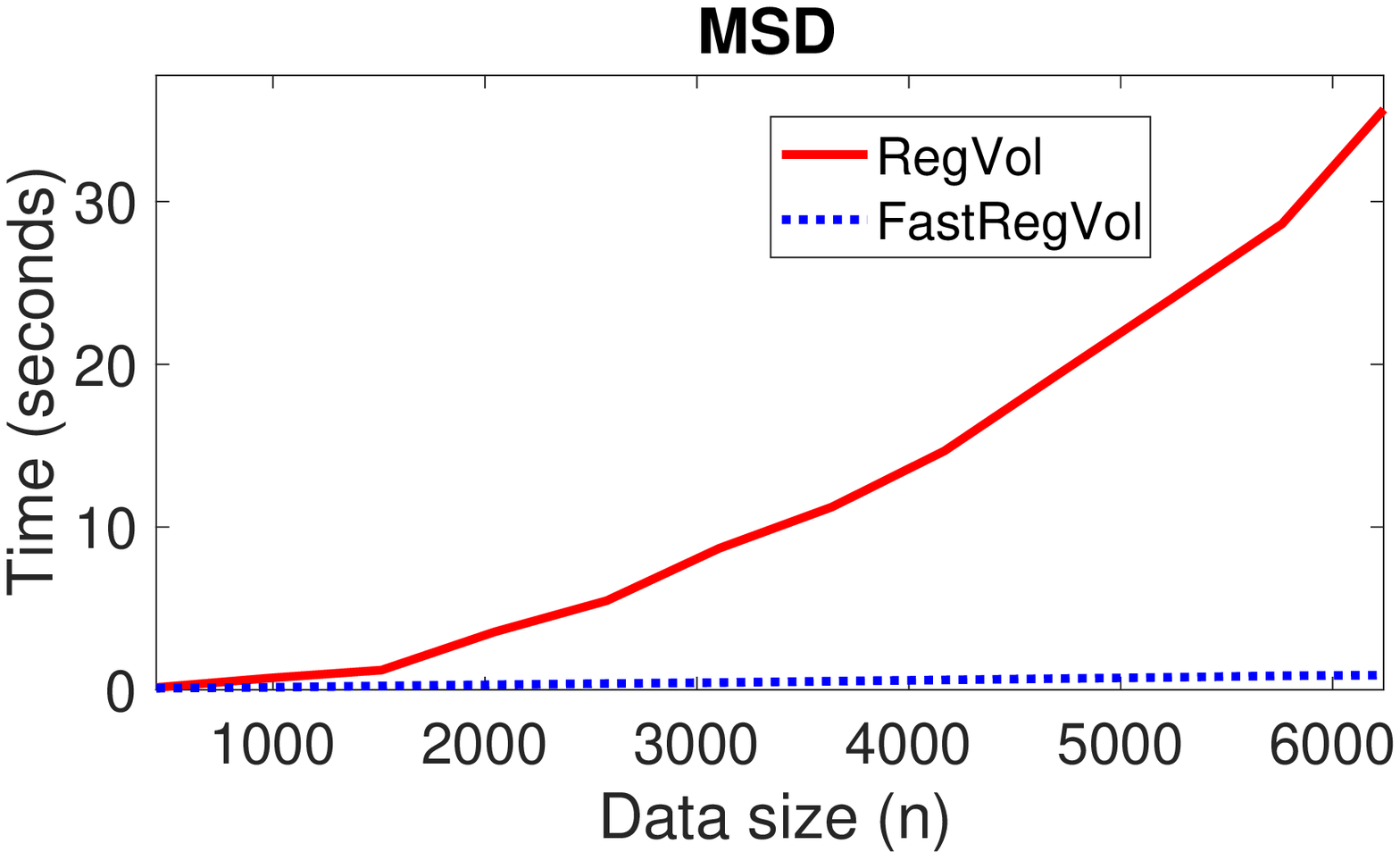}
\includegraphics[width=0.25\textwidth]{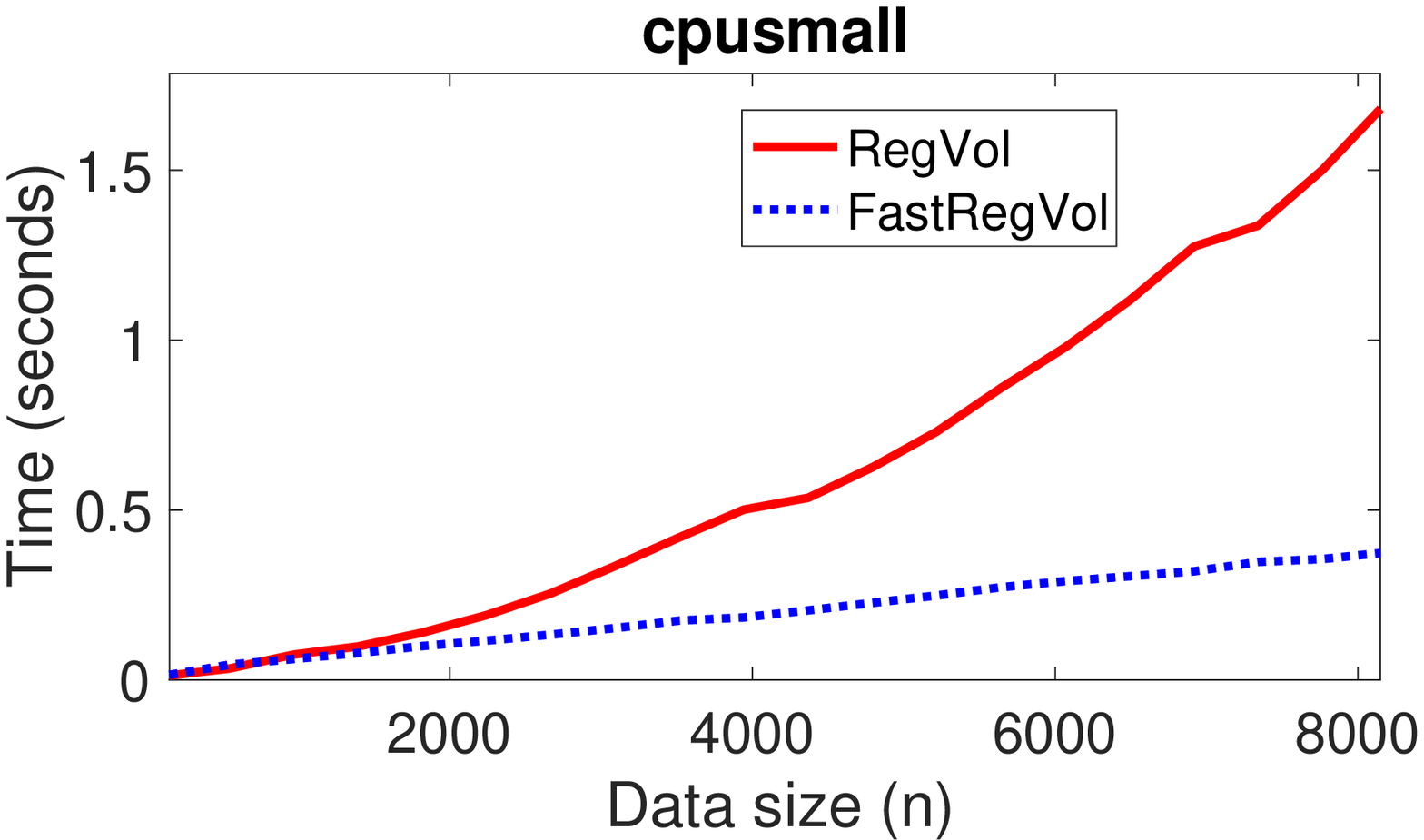}\nobreak
\includegraphics[width=0.25\textwidth]{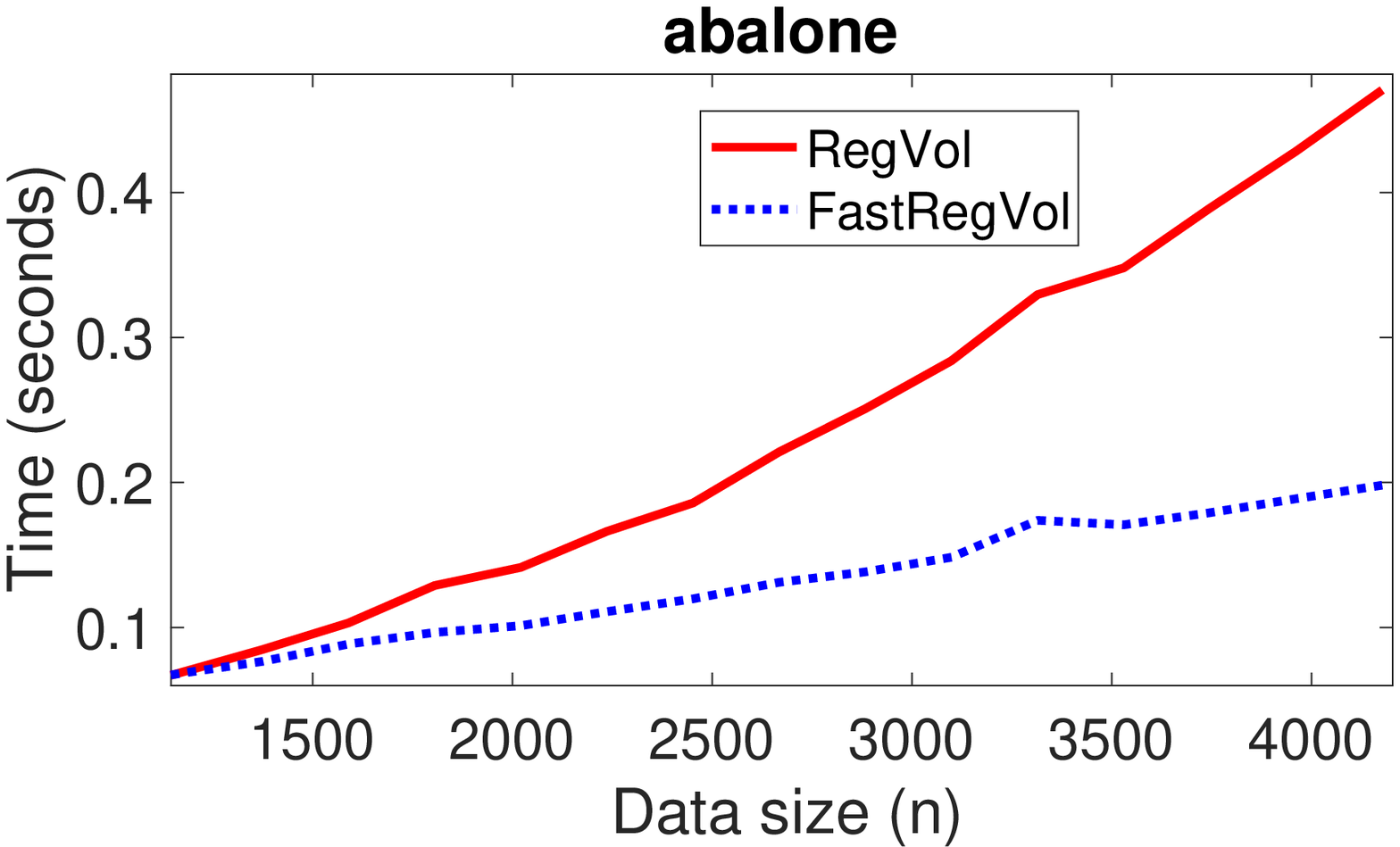}
\caption{Comparison of runtime between {\lazy} (see Alg. \ref{alg:lazyvolsamp})
  and {\volsamp} \cite{unbiased-estimates} on four datasets, with the
  methods ran on data subsets of varying size (n).}
\label{fig:time-plots}
\end{figure}

\vspace{-4mm}
In Figure \ref{fig:time-plots} we plot the runtime against varying
values of $n$ (using portions of the datasets), to
compare how {\lazy} and {\volsamp} scale with
respect to the datasize. We observe that unlike {\volsamp}, our new algorithm
exhibits linear dependence on $n$, thus it is much better suited for
running on large datasets. 

\subsection{Subset selection for ridge regression}

We applied volume sampling to the task of subset selection for
linear regression, by evaluating the subsampled ridge estimator $\of{\wbh_\lambda^*}S$ using
the total loss over the full dataset:
\vspace{-3mm}
\[L(\of{\wbh_\lambda^*}S) \defeq \frac{1}{n}\|\X^\top\of{\wbh_\lambda^*}S-\y\|^2.\]
\vspace{-.6cm}

\begin{figure}
\includegraphics[width=0.25\textwidth]{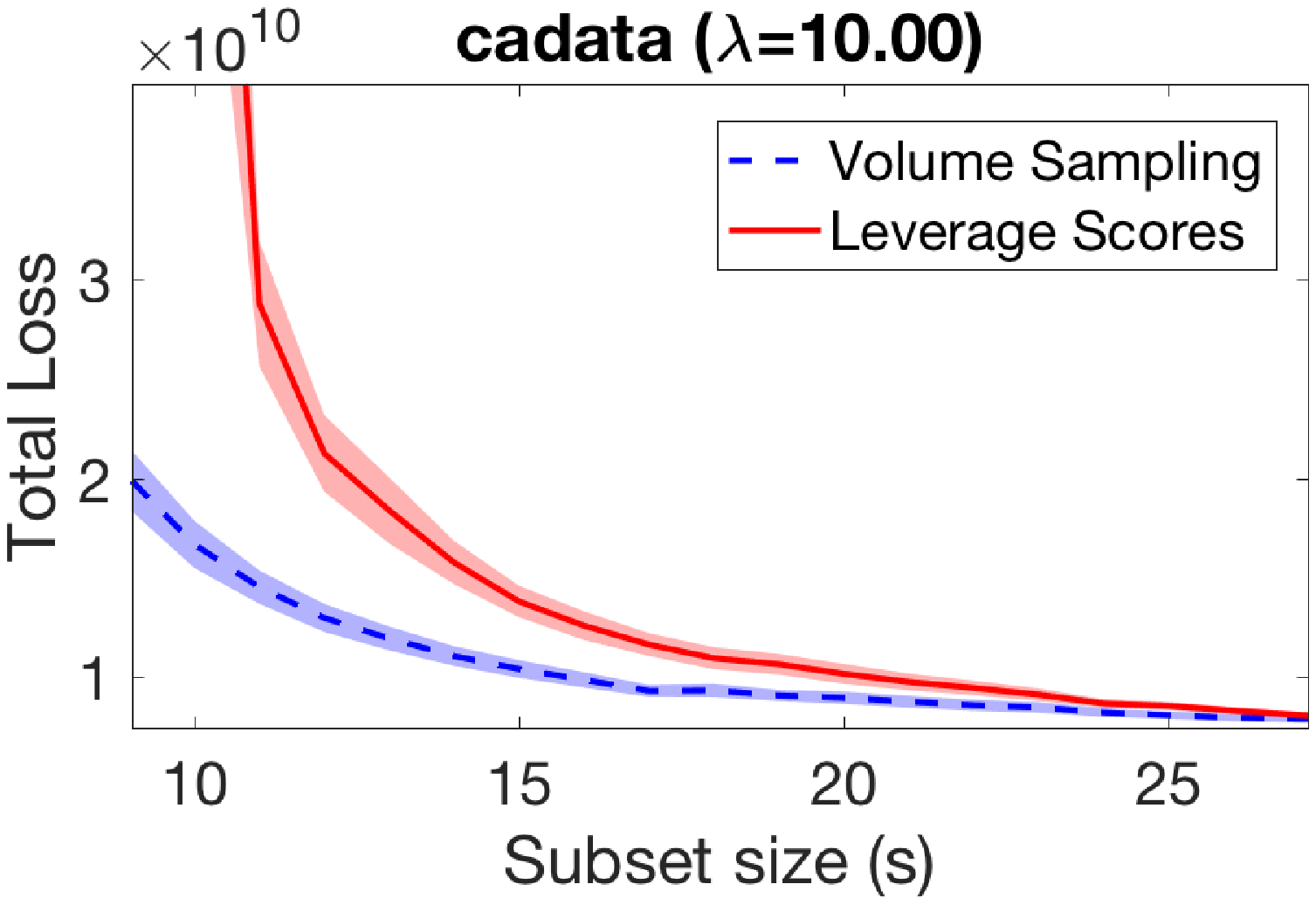}\nobreak
\includegraphics[width=0.25\textwidth]{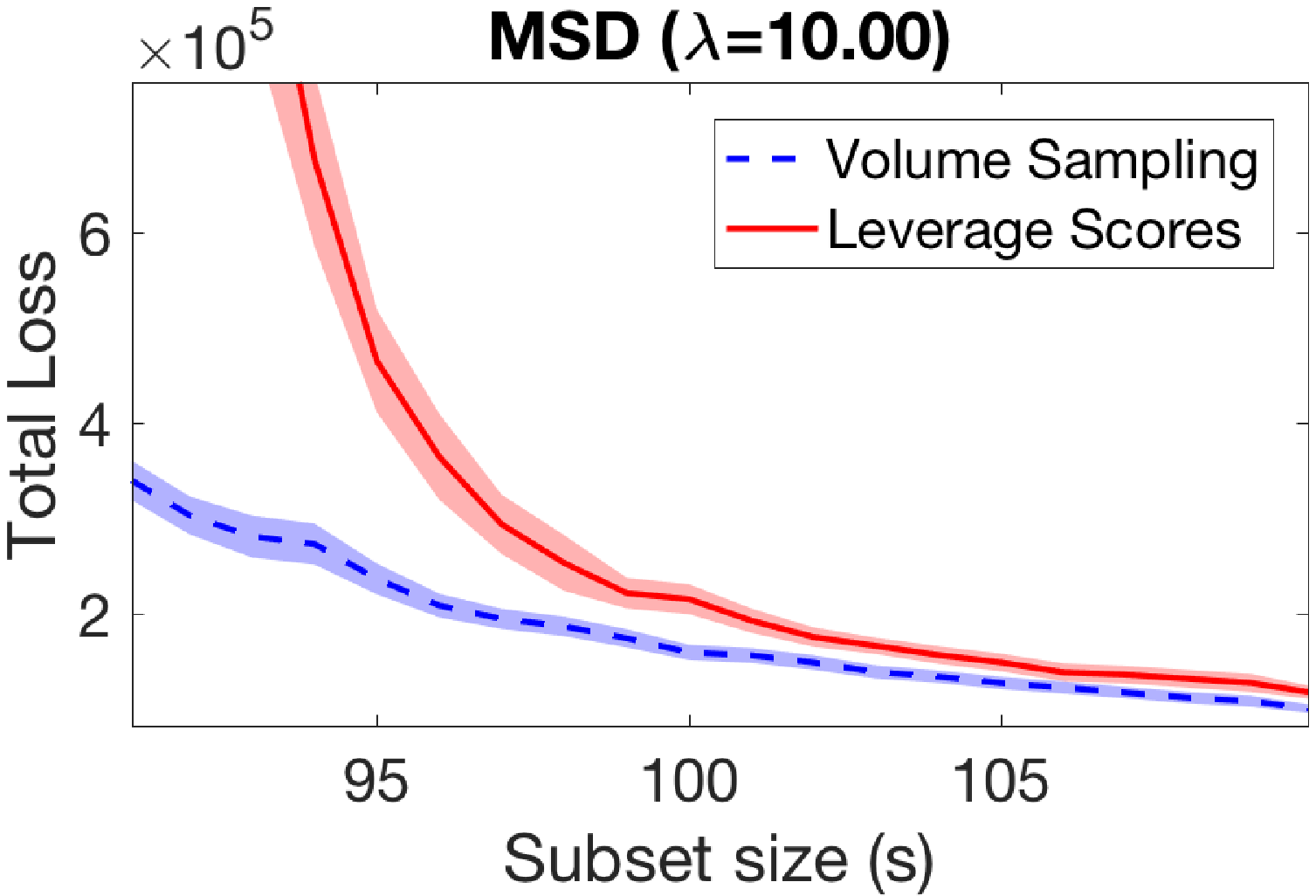}
\includegraphics[width=0.25\textwidth]{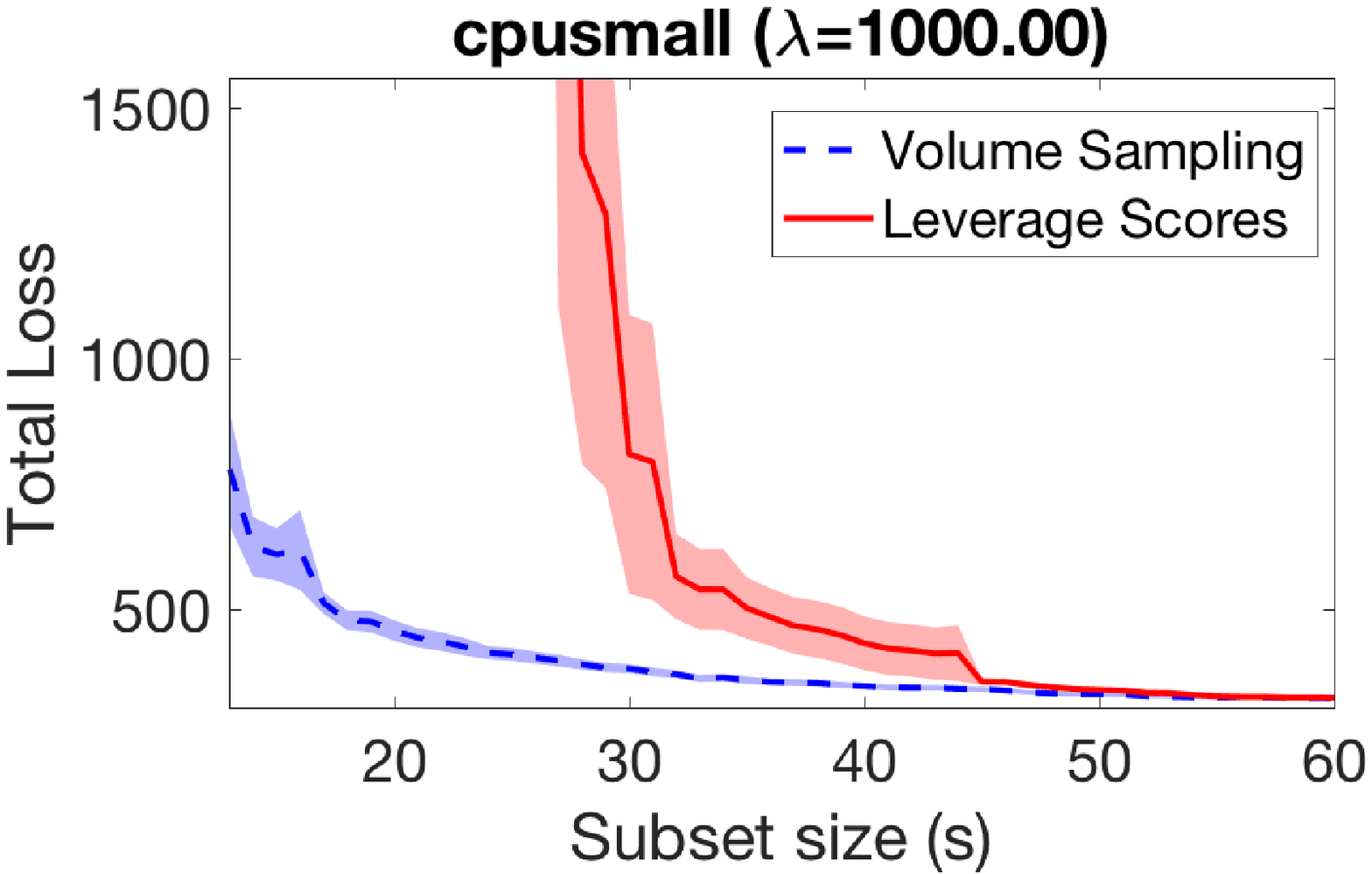}\nobreak
\includegraphics[width=0.25\textwidth]{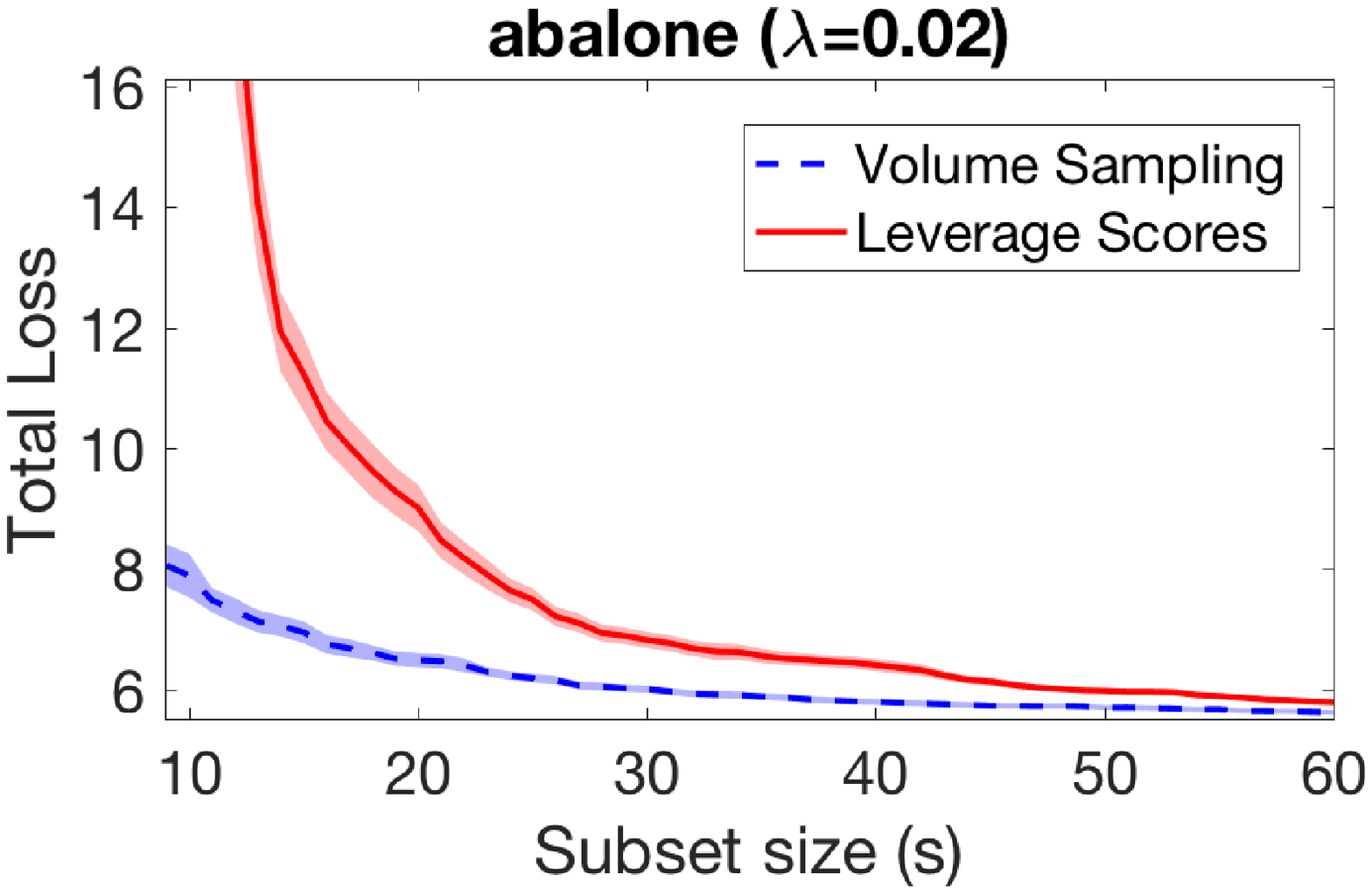}
\caption{Comparison of loss of the subsampled ridge estimator when
  using regularized volume sampling vs using leverage score sampling
  on four datasets.}
\label{fig:predictions}
\end{figure}

We computed $L(\of{\wbh_\lambda^*}S)$ for a range of
subset sizes and values of $\lambda$, when the subsets are sampled
according to $\lambda$-regularized volume sampling%
\footnote{Our experiments suggest that using the same
  $\lambda$ for sampling and for computing the ridge estimator works
  best.} 
%Same $\lambda$ was used for sampling and the ridge estimator.
 and leverage score sampling. The results were averaged over 20 runs of each
experiment. For clarity, Figure \ref{fig:predictions}
shows the results only with one value of $\lambda$ for each
dataset, chosen so that the subsampled ridge estimator performed best
(on average over all samples of preselected size $s$). 
% Figure \ref{fig:predictions} plots the total loss
% <<<<<<< HEAD
% $L(\of{\wbh_\lambda^*}S)$ for a tuned value of $\lambda$,
% against the size of subset $S$
% %for a range of $\lambda$'s, 
% sampled according to
% $\lambda$-regularized volume sampling (with the same value of $\lambda$) and leverage score sampling. Note that leverage score
% sampled subproblems require appropriate rescaling of the instances
% before solving for $\of{\wbh_\lambda^*}S$ (see
% \cite{randomized-matrix-algorithms} for details), which we applied in our
% experiments, whereas volume sampling does not require any rescaling.
% =======
% $L(\of{\wbh_\lambda^*}S)$ against the size of subset $S$ sampled according to
% $\lambda$-regularized volume sampling (with the same value of
% $\lambda$) and leverage score sampling. 
% For each dataset and sampling method, 
% we show the $L(\of{\wbh_\lambda^*}S)$ curve for the ???best???
% choice of parameter $\lambda$.
Note that for leverage scores
we did the appropriate rescaling of the instances
before solving for $\of{\wbh_\lambda^*}S$ 
for the sampled subproblems
(see \cite{randomized-matrix-algorithms} for details). 
Volume sampling does not require any rescaling.
% >>>>>>> 5adff596745400ee03c6c0fa73430fe50a0ed079
%The shown plots are averaged over 20 runs of the experiment.
The results on all datasets show that when only a small number of
labels $s$ is obtainable, then regularized volume sampling offers better estimators than
leverage score sampling (as predicted by Theorems \ref{thm:ridge} and \ref{thm:lb-all}). 
% The lower-bound from Theorem \ref{thm:lb-all}
% part 2 can be observed for dataset cpusmall, where $d=12$ and $d\log d\approx 30$),
% such differences can occur in practice.

\section{Conclusions}
\label{sec:conclusions}
We proposed a sampling procedure called regularized volume sampling,
which offers near-optimal statistical guarantees for subsampled ridge
estimators. We also gave a new algorithm for volume sampling
which is essentially as efficient as i.i.d. leverage score
sampling.

%\subsubsection*{References}
\bibliographystyle{plain}
\ifisarxiv
\bibliography{ridge}
\else
\bibliography{pap}

\begin{thebibliography}{10}

\bibitem{ridge-leverage-scores}
Ahmed~El Alaoui and Michael~W. Mahoney.
\newblock Fast randomized kernel ridge regression with statistical guarantees.
\newblock In {\em Proceedings of the 28th International Conference on Neural
  Information Processing Systems}, pages 775--783, Montreal, Canada, December
  2015.

\bibitem{tractable-experimental-design}
Zeyuan Allen-Zhu, Yuanzhi Li, Aarti Singh, and Yining Wang.
\newblock Near-optimal design of experiments via regret minimization.
\newblock In {\em Proceedings of the 34th International Conference on Machine
  Learning}, volume~70 of {\em Proceedings of Machine Learning Research}, pages
  126--135, Sydney, Australia, August 2017.

\bibitem{avron-boutsidis13}
Haim Avron and Christos Boutsidis.
\newblock Faster subset selection for matrices and applications.
\newblock {\em SIAM Journal on Matrix Analysis and Applications},
  34(4):1464--1499, 2013.

\bibitem{ridge-uniform}
Francis Bach.
\newblock Sharp analysis of low-rank kernel matrix approximations.
\newblock In {\em Proceedings of the 26th Annual Conference on Learning
  Theory}, volume~30 of {\em Proceedings of Machine Learning Research}, pages
  185--209, Princeton, NJ, USA, June 2013.

\bibitem{unbiased-estimates}
Micha{\l} Derezi\'{n}ski and Manfred~K Warmuth.
\newblock Unbiased estimates for linear regression via volume sampling.
\newblock In {\em Advances in Neural Information Processing Systems 30}, pages
  3087--3096, Long Beach, CA, USA, December 2017.

\bibitem{efficient-volume-sampling}
Amit Deshpande and Luis Rademacher.
\newblock Efficient volume sampling for row/column subset selection.
\newblock In {\em Proceedings of the 2010 IEEE 51st Annual Symposium on
  Foundations of Computer Science}, pages 329--338, Las Vegas, USA, October
  2010.

\bibitem{pca-volume-sampling}
Amit Deshpande, Luis Rademacher, Santosh Vempala, and Grant Wang.
\newblock Matrix approximation and projective clustering via volume sampling.
\newblock In {\em Proceedings of the Seventeenth Annual ACM-SIAM Symposium on
  Discrete Algorithm}, pages 1117--1126, Miami, FL, USA, January 2006.

\bibitem{fast-leverage-scores}
Petros Drineas, Malik Magdon-Ismail, Michael~W. Mahoney, and David~P. Woodruff.
\newblock Fast approximation of matrix coherence and statistical leverage.
\newblock {\em J. Mach. Learn. Res.}, 13(1):3475--3506, December 2012.

\bibitem{optimal-design-book}
Valerii~V. Fedorov, William~J. Studden, and E.~M. Klimko, editors.
\newblock {\em Theory of optimal experiments}.
\newblock Probability and mathematical statistics. Academic Press, New York,
  NY, USA, 1972.

\bibitem{dpp-shopping}
Mike Gartrell, Ulrich Paquet, and Noam Koenigstein.
\newblock Bayesian low-rank determinantal point processes.
\newblock In {\em Proceedings of the 10th ACM Conference on Recommender
  Systems}, pages 349--356, Boston, MA, USA, September 2016.

\bibitem{more-efficient-volume-sampling}
Venkatesan Guruswami and Ali~K. Sinop.
\newblock Optimal column-based low-rank matrix reconstruction.
\newblock In {\em Proceedings of the Twenty-third Annual ACM-SIAM Symposium on
  Discrete Algorithms}, pages 1207--1214, Kyoto, Japan, January 2012.

\bibitem{geometric-tail-bounds}
Svante Janson.
\newblock Tail bounds for sums of geometric and exponential variables.
\newblock {\em Statistics and Probability Letters}, 135:1 -- 6, 2018.

\bibitem{k-dpp}
Alex Kulesza and Ben Taskar.
\newblock {k-DPPs: Fixed-Size Determinantal Point Processes}.
\newblock In {\em {Proceedings of the 28th International Conference on Machine
  Learning}}, pages 1193--1200, Bellevue, WA, USA, June 2011.

\bibitem{dpp}
Alex Kulesza and Ben Taskar.
\newblock {\em Determinantal Point Processes for Machine Learning}.
\newblock Now Publishers Inc., Hanover, MA, USA, 2012.

\bibitem{dual-volume-sampling}
Chengtao Li, Stefanie Jegelka, and Suvrit Sra.
\newblock Polynomial time algorithms for dual volume sampling.
\newblock In {\em Advances in Neural Information Processing Systems 30}, pages
  5045--5054, Long Beach, CA, USA, December 2017.

\bibitem{uci-repository}
M.~Lichman.
\newblock {UCI} machine learning repository, 2013.

\bibitem{randomized-matrix-algorithms}
Michael~W. Mahoney.
\newblock Randomized algorithms for matrices and data.
\newblock {\em Found. Trends Mach. Learn.}, 3(2):123--224, February 2011.

\bibitem{pool-based-active-learning-regression}
Masashi Sugiyama and Shinichi Nakajima.
\newblock Pool-based active learning in approximate linear regression.
\newblock {\em Mach. Learn.}, 75(3):249--274, June 2009.

\end{thebibliography}
\fi

\end{document}